\documentclass[pdflatex,sn-mathphys-num]{sn-jnl}

\usepackage{graphicx}%
\usepackage{multirow}%
\usepackage{amsmath,amssymb,amsfonts}%
\usepackage{amsthm}%
\usepackage{mathrsfs}%
\usepackage[title]{appendix}%
\usepackage{xcolor}%
\usepackage{textcomp}%
\usepackage{manyfoot}%
\usepackage{booktabs}%
\usepackage{algorithm}%
\usepackage{algorithmicx}%
\usepackage{algpseudocode}%
\usepackage{listings}%

\theoremstyle{thmstyleone}%
\theoremstyle{thmstyletwo}%

\theoremstyle{thmstylethree}%

\begin{document}

\title[Article Title]{ Integrated Oculomics and Lipidomics Reveal Microvascular–Metabolic Signatures Associated with Cardiovascular Health in a Healthy Cohort}


\author*[1]{\fnm{Inamullah}} \email{i1n23@soton.ac.uk}

\author[1]{\fnm{Ernesto Elias} \sur{Vidal Rosas}}\email{E.E.Vidal-Rosas@soton.ac.uk}
\author[2]{\fnm{Imran} \sur{Razzak}}\email{Imran.Razzak@mbzuai.ac.ae}

\author[1]{\fnm{Shoaib} \sur{Jameel}}\email{M.S.Jameel@southampton.ac.uk}

\affil*[1]{\orgdiv{Electronics and Computer Science}, \orgname{University of Southampton}, \orgaddress{\street{}, \city{Southampton}, \postcode{SO18 1PB}, \state{Hampshire}, \country{United Kingdom}}}

\affil[2]{\orgdiv{Department of Computational Biology}, \orgname{Mohamed bin Zayed University of Artificial Intelligence}, \orgaddress{\state{Abu Dhabi}, \country{United Arab Emirates}}}

\abstract{Cardiovascular disease (CVD) remains the leading global cause of mortality, yet current risk stratification methods often fail to detect early, subclinical changes. Previous studies have generally not integrated retinal microvasculature characteristics with comprehensive serum lipidomic profiles as potential indicators of CVD risk. In this study, an innovative imaging-omics framework was introduced, combining retinal microvascular traits—derived through deep learning-based image processing—with serum lipidomic data to highlight asymptomatic biomarkers of cardiovascular risk beyond the conventional lipid panel. This represents the first large-scale, covariate-adjusted and stratified correlation analysis conducted in a healthy population, which is essential for identifying early indicators of disease.
Retinal phenotypes were quantified using automated image analysis tools, while serum lipid profiling was performed by Ultra-High-Performance Liquid Chromatography Electrospray ionization High-resolution mass spectrometry (UHPLC-ESI-HRMS). Strong, age- and sex-independent correlations were established, particularly between average artery width, vessel density, and lipid subclasses such as triacylglycerols (TAGs), diacylglycerols (DAGs), and ceramides (Cers). These associations suggest a converging mechanism of microvascular remodeling under metabolic stress. By linking detailed vascular structural phenotypes to specific lipid species, this study fills a critical gap in the understanding of early CVD pathogenesis. This integration not only offers a novel perspective on microvascular–metabolic associations but also presents a significant opportunity for the identification of robust, non-invasive biomarkers. Ultimately, these findings may support improved early detection, targeted prevention, and personalized approaches in cardiovascular healthcare.}


\maketitle

\section{Introduction}\label{sec1}
CVD remains the leading cause of mortality and morbidity worldwide, accounting for nearly 18 million deaths annually \cite{bib1}. While current risk-stratification algorithms rely on traditional clinical factors such as blood pressure, age, smoking history, and standard lipid panels, major adverse cardiovascular events often stem from complex interactions between systemic metabolic dysregulation and subclinical microvascular alterations that can occur before overt disease symptoms emerge \cite{bib2-rudnicka2022artificial}. This highlights the critical need for advanced, non-invasive approaches to identify individuals at elevated CVD risk earlier.

The human eye offers a unique, non-invasive window into the body's microvasculature. Examining the retinal microvasculature, which includes arterioles and venules, can reveal early signs of circulatory-related pathological processes. The extensive microvascular network branching throughout the body significantly affects organ health and disease susceptibility, and abnormalities in this network can lead to severe systemic and ocular conditions \cite{bib3-hanssen2022retinal}. The anatomical, physiological, and embryological similarities between the retinal microvasculature and other organ systems, particularly the coronary and cerebral microcirculations, make the eye a powerful predictive biomarker for systemic health \cite{bib4-zhu2025oculomics}. Changes in retinal vessel morphology, for instance, reflect systemic vascular ageing and endothelial dysfunction—key mechanisms underlying CVD pathogenesis \cite{bib3-hanssen2022retinal}. Consequently, understanding retinal microvascular dynamics holds considerable potential for advancing early detection and improving risk stratification in cardiovascular disease.

Recent advancements in automated artificial intelligence (AI) software, big data analytics, and widespread ophthalmic imaging have enabled detailed exploration of ocular and systemic health, a field now termed ``oculomics'' \cite{bib4-zhu2025oculomics}. The accessibility of retinal imaging and strong public participation in eye screening programs further support its use as a non-invasive tool for cardiovascular disease assessment \cite{bib5-wagner2020insights}. For example, studies have utilised software to predict CVD risk from fundus images \cite{bib6-arnould2021prediction,bib7-kim2020effects}, and several have reported significant associations between fundus retinal images and a range of CVD-related risk factors \cite{bib8-chang2020association, bib9-aschauer2021identification,bib10-nusinovici2022retinal,bib11-zekavat2022deep}. While manual interpretation of these images is time-consuming, recent AI methodologies \cite{bib12} enable scalable, automated extraction of vascular phenotypes. Moreover, research by Rudnicka et al. \cite{bib2-rudnicka2022artificial} demonstrated that retinal vasculometry alone can serve as a circulatory biomarker for assessing stroke, cardiovascular mortality, and myocardial infarction risks, even without traditional clinical measurements.

In parallel with microvascular insights, the serum metabolome, particularly its lipidomic profile, offers a comprehensive view of systemic metabolic health. Lipidomics, a subfield of metabolomics, involves profiling cellular lipids crucial for hormonal regulation, energy storage, membrane structure, and signalling pathways. These functions position circulating lipids as key players in the pathophysiology of age-related diseases, including cardiovascular disease and stroke \cite{bib13-qin2020insights}. Unlike conventional lipid measures (e.g., total cholesterol, Low density Lipoprotein cholesterol (LDL-C), High density Lipoprotein cholesterol (HDL-C) ) that offer limited granularity, advanced lipidomics platforms, such as Ultra-High-
Performance Liquid Chromatography Electrospray ionization High-resolution mass spectrometry (UHPLC-ESI-HRMS), enable the quantification of hundreds to thousands of lipid species across diverse structural classes \cite{bib14-han2016lipidomics}.

Lipidomics encompasses a wide array of molecular species that can be broadly grouped into biologically distinct classes. These include glycerolipids—such as triacylglycerols (TAGs) and diacylglycerols (DAGs)—which are central to energy storage and metabolic signaling; sphingolipids—including ceramides (Cers) and sphingomyelins (SM)—which are involved in membrane structure, inflammation, and apoptosis; and glycerophospholipids, comprising phosphatidylcholines (PC), lysophospholipids (lysoPCs), and phosphatidylethanolamines (PE), which play key roles in membrane dynamics and lipid transport. Cholesteryl esters (CE) and free fatty acids (FFA) are also frequently profiled in lipidomics studies due to their relevance in lipid metabolism. Dysregulation in these lipid classes has been linked to processes such as atherosclerosis, insulin resistance, oxidative stress, and vascular dysfunction—all of which are central to the pathophysiology of cardiovascular disease \cite{bib15-meikle2014lipidomics, bib16-stegemann2014lipidomics,bib17-chen2023longitudinal}. Studies have further highlighted the prognostic value of lipidomic biomarkers, with plasma ceramides, for example, predicting cardiovascular death independently of cholesterol measures \cite{bib18-meikle2011plasma, bib19-laaksonen2016plasma}.

Numerous studies have explored the association between fundus characteristics, such as retinal vessel caliber, and systemic diseases such as hypertension, dyslipidemia, obesity, smoking, and glycemic control \cite{bib20-tapp2019associations,bib21-ikram2006retinal,bib22-lemmens2022age}, underscoring retinal imaging's potential as a systemic health biomarker. However, much of the existing research is constrained by small sample sizes, a narrow focus on risk factors, and insufficient integration of multimodal data, including imaging, omics, and electronic health records. Many studies also rely on heterogeneous cohorts with pre-existing conditions, which can obscure signals relevant to early-stage disease detection. Critically, the lack of data from healthy participants limits the ability to identify biomarkers predictive of disease onset.

Despite the biological plausibility linking retinal microvascular features and systemic lipid abnormalities, few large-scale investigations have combined retinal imaging with high-dimensional lipidomics to explore their association with cardiovascular risk. Prior work has often focused on traditional lipid metrics, addressed diseases unrelated to CVD, or examined localised disease manifestations. For example, while some studies have shown that lipid profiles can predict early retinal hard exudates in diabetic retinopathy \cite{bib23-sasaki2013quantitative, bib24-shen2023novel}, and deep learning models applied to fundus images have predicted cardiovascular risk factors \cite{bib2-rudnicka2022artificial}, the mechanistic links between retinal features and specific metabolic intermediates, such as lipid subclasses, remain largely underexplored. Furthermore, efforts to combine imaging and omics have been limited in scope or focused on diseased populations, with image-based prediction models often lacking mechanistic interpretability and trained on mixed clinical datasets \cite{bib25-poplin2018prediction, bib26-zhang2021deep}.

Building upon the insights above, our study makes several significant contributions to the field of cardiovascular risk assessment. To the best of our knowledge, this is the first study to conduct a comprehensive, adjusted, and stratified correlation analysis between high-resolution serum lipidomic profiles and deep-learning-derived retinal microvascular features within a large, rigorously selected healthy cohort. This deliberate focus on healthy participants is crucial for identifying early, pre-symptomatic biomarkers of disease onset, which is often obscured in studies relying on heterogeneous cohorts with pre-existing conditions.

Our methodological choices are driven by the need for both breadth and depth in understanding cardiovasculometabolic health. We leverage deep learning for retinal microvascular feature extraction because it enables scalable, automated, and precise quantification of subtle vascular phenotypes that are highly indicative of systemic microcirculatory health \cite{bib12}. This approach overcomes the limitations of manual interpretation, which is time-consuming and prone to inter-observer variability. Simultaneously, we employ ultra-high-performance liquid chromatography coupled with high-resolution mass spectrometry for serum lipidomics profiling \cite{bib14-han2016lipidomics}. This advanced platform allows for the quantification of hundreds to thousands of distinct lipid species, providing useful granularity compared to traditional lipid panels. This comprehensive lipidomic analysis is essential for uncovering specific lipid subclasses and pathways implicated in early atherosclerosis, insulin resistance, and vascular dysfunction, which are otherwise over looked by conventional methods \cite{bib15-meikle2014lipidomics,bib16-stegemann2014lipidomics}. 

By directly linking granular vascular structural information with detailed molecular lipid species, our work bridges a critical gap in understanding the intricate interplay between metabolic dysregulation and microvascular alterations. This novel multimodal integration of lipidomics with advanced retinal imaging represents a powerful and hitherto underutilised opportunity to discover robust, non-invasive biomarkers for cardiovascular and metabolic risk stratification. Ultimately, such comprehensive, data-driven approaches are essential for moving beyond conventional risk factors, enabling earlier detection, more effective preventive strategies, and truly personalised healthcare in cardiovascular medicine. This work paves the way for new avenues in biomarker discovery and the development of refined metabolic–vascular phenotyping strategies.

\begin{figure}
\centering

\begin{minipage}[b]{1.0\textwidth}
  \centering
  \includegraphics[width=\linewidth]{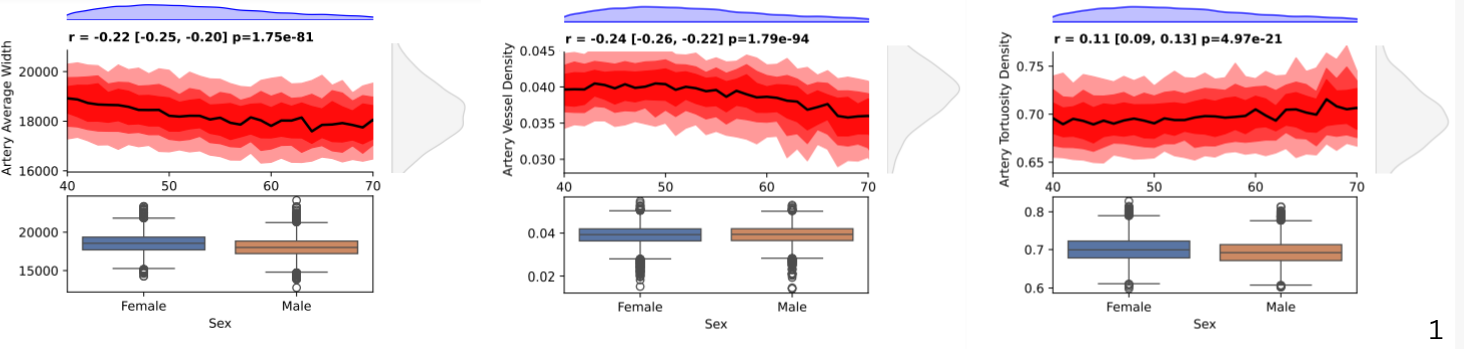}
  \\ \small (a)
\end{minipage}
\hfill
\begin{minipage}[b]{1.0\textwidth}
  \centering
  \includegraphics[width=\linewidth]{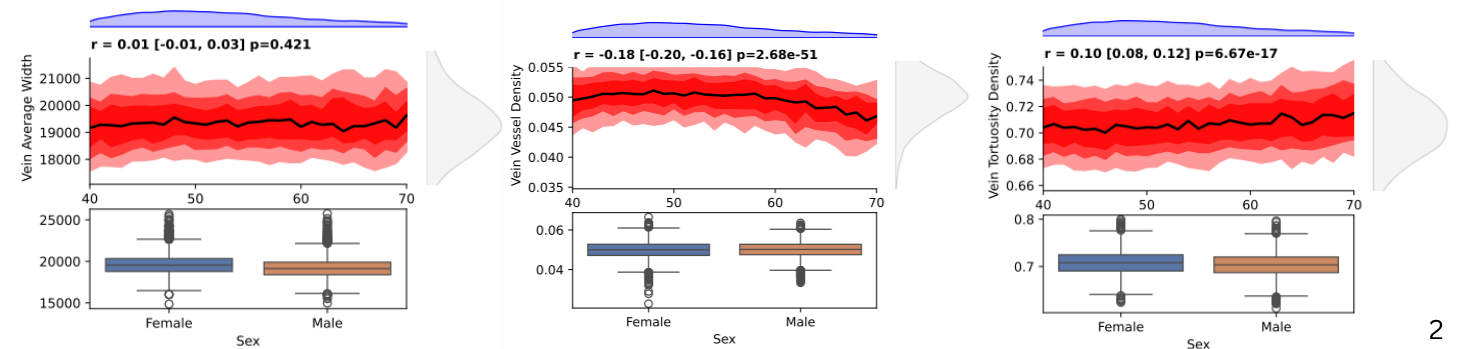}
  \\ \small (b)
\end{minipage}
\caption{The 18 retinal microvascular vessels were measured as covariates, with adjustments made for age and sex, while six key features from both veins and arteries were selected for simplicity. The Benjamini-Hochberg (BH) false discovery rate (FDR) correction was applied using the Python Pingouin package to account for multiple hypotheses. Additionally, a 95\% confidence interval (CI) was incorporated, detailing both the lower and upper bounds.}
\label{fig:1}
\end{figure}
\pagebreak

\section{Results}\label{sec2}

CVD remains the world’s leading cause of fatality. Traditional diagnostics often lack understanding at initial disease stages, drawing attention to the need for novel biomarkers \cite{bib27-zhang2024recent}. Retinal imaging is indispensable for capturing the fine-grained patterns present in the microcirculation of blood vessels in human health. This provides a proxy calculator for calculating the vascular risk, facilitating the estimation of different cardiovascular events and mortality\cite{bib28-de2014fundus}. The authors of this research build upon the work presented in \cite{bib29-reicher2024phenome}, aiming to deepen the understanding of the limitations in unravelling lipid profiling patterns during the menopausal transition in relation to CVD risk. This investigation seeks to identify specific lipid profiles that may serve as potential biomarkers. In a related study \cite{bib30-shapira2024unveiling}, the authors adjusted for age and sex and examined six fundus features, which revealed distinct clinical characteristics. However, this study is hindered by the absence of subspecies analysis and the capability to investigate lipids at the microlevel, which is essential for effective CVD risk stratification that is often confined to broader clinical covariates.

\begin{figure}[!htb]
  \centering
  \includegraphics[width=1.0\textwidth]{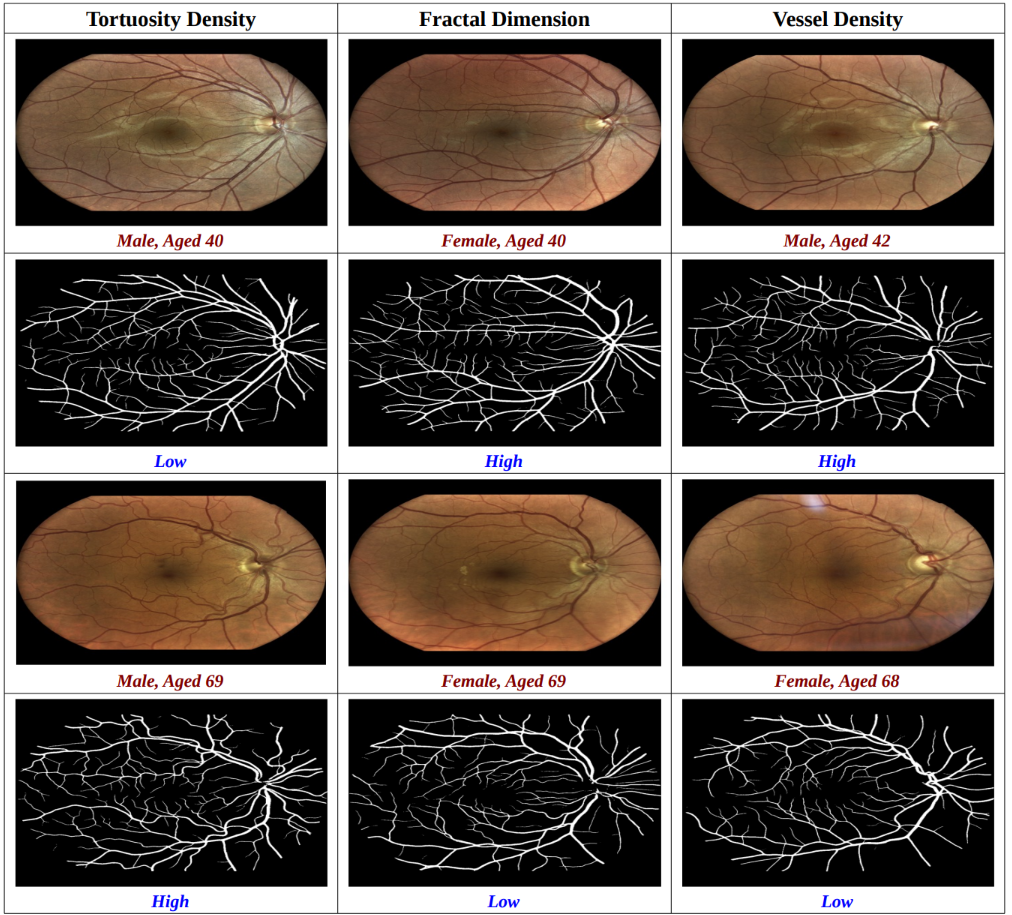}
  \caption{\textbf{Visualization of retinal microvascular features across age and sex using fundus image segmentation.}
Each column represents a distinct microvascular trait—\textit{tortuosity density}, \textit{fractal dimension}, and \textit{vessel density}. For each feature, younger participants are shown in the top two rows and older participants in the bottom two rows. Raw fundus images are followed by their corresponding segmented vessel maps. Annotations indicate age, sex, and whether the feature expression is high or low. The images illustrate that tortuosity increases with age, while fractal dimension and vessel density tend to decline, revealing vascular remodelling patterns associated with healthy aging.}
  \label{fig:2}
\end{figure}
\clearpage

\subsection{Population Characteristics, Age and sex adjustment}
Fundus images were processed using Automorph tools \cite{bib12}, resulting in 7,068 validated images with baseline demographic data. This dataset demonstrated a balanced sex distribution, with a mean age consistent among the lipidomics and fundus sub-cohorts (mean age for fundus: 52.64 ± 7.87 years; mean age for lipids: 51.76 ± 8.09 years). The male and female counts in the lipidomics and fundus datasets were 46.6\%, 48.7\%, and 53.4\%, 51.3\%, respectively.

Key microvascular characteristics of the fundus fell within expected ranges, with an average artery width of 18,305 ± 1,287 pixels and a vessel density of approximately 0.039 ± 0.0045 (see Table 1). The initial analysis evaluated 18 fundus traits using Pearson correlation, treating sex and age as covariates. Corrections were calculated using the Pingouin package, which served as an independent variable across all features. For each of the 18 fundus traits, we computed the correlation coefficient (r), p-value, and 95\% CI. The FDR-BH correction  was applied to identify statistically significant traits.

A multi-panel visualisation was then created for all 18 features, illustrating the kernel density estimation (KDE) distribution of age, percentile bands of fundus traits across age, and sex-specific boxplots. Features with an absolute correlation coefficient $|r|$ \(\geq\) $0.1$ and FDR-adjusted p-values $<$ $0.05$ were flagged for further analysis. After accounting for multiple testing, the number of retinal features was reduced from 18 to 15, and six key features were identified. These features were corroborated through pairwise Pearson correlation and hierarchical clustering.

\subsubsection{Association of Retinal characteristic with age and sex}
The six key retinal features exhibited different progression patterns related to age and sex. Specifically, the average width and tortuosity density for both arteries and veins displayed similar patterns regarding gender differences. Additionally, while the vessel density for arteries and veins was similar across sexes, it exhibited distinct progression trends with age.

This study revealed that statistically significant correlations with age, exceeding an absolute value of 0.1, were demonstrated by five out of the six selected vascular features and more than 10 other features. The Fig \ref{fig:2} derived from the fundus dataset provides a compelling rationale for understanding late-life diseases in asymptomatic patients who exhibit manifestations without clear signs. This microvasculature warrants careful examination. Tortuosity density is higher in older adults and lower in younger individuals, whereas vessel complexity and the presence of dense vessels are elevated in younger age groups and diminished in older ones. This suggests the importance of early identification and clinical diagnosis to facilitate timely inspection and intervention for various ocular and non-ocular diseases. Such insights offer a significant opportunity to explore these issues from multiple perspectives and a comprehensive viewpoint. To investigate and to study critically both the arterial and venous vascular traits, looking at the age-related patterns, follows. This reveals a reliable trend among the several features. The artery average width illustrates a significant negative association with the progression of age  (r = –0.22, 95\% CI: –0.25 to –0.20, p = 1.75e 81), as an indication of nicking of the artery with the age factors. Likewise, artery vessel density decreases with age ( (r = –0.24, 95\% CI: –0.26 to –0.22, p = 1.79e–94)), pointing to the rarefaction of the arterial network over time.  

In contrast, the atery tortuosity density grows with age (r = 0.11, 95\% CI: 0.09 to 0.13, p = 4.97e-21), demonstrating the compensatory vascular remodelling or age-associated structural changes. Discussing the venule pattern, they exhibit distinct patterns compared to arterioles.The venous measures, such as the vein vessel shrinking with age (r = –0.18, 95\% CI: –0.20 to –0.16, p = 2.68e–51), and vein tortusity density displaying a weak however a significant increase (r = 0.10, 95\% CI: 0.08 to 0.12, p = 6.67e–17). The last in the six vascular measures is the vein average width, which shows no significant correlation with age (r = 0.01, p = 0.421), suggesting the stability across the studies' age range in terms of this feature. 

Sex-based differences were observed across several retinal vascular features. Females exhibited slightly greater average artery and vein widths compared to males, as shown in the KDE plots and the boxplots in Fig \ref{fig:1}. Tortuosity density was also consistently higher in females for both arteries and veins, indicating more intricate vascular geometry. While artery vessel density declined with age in both sexes, females maintained marginally higher values, contrasting with earlier studies that reported no sex difference in vessel density.

These results emphasize the value of stratifying analyses by sex, as the patterns of vascular ageing—though directionally similar—varied in magnitude and rate between males and females. Such differences highlight the need to account for demographic covariates in retinal biomarker studies and reinforce the importance of sex-specific vascular phenotyping in oculomics research.
\pagebreak

\begin{figure}[!htb]
  \centering
  \includegraphics[width=1.0\textwidth]{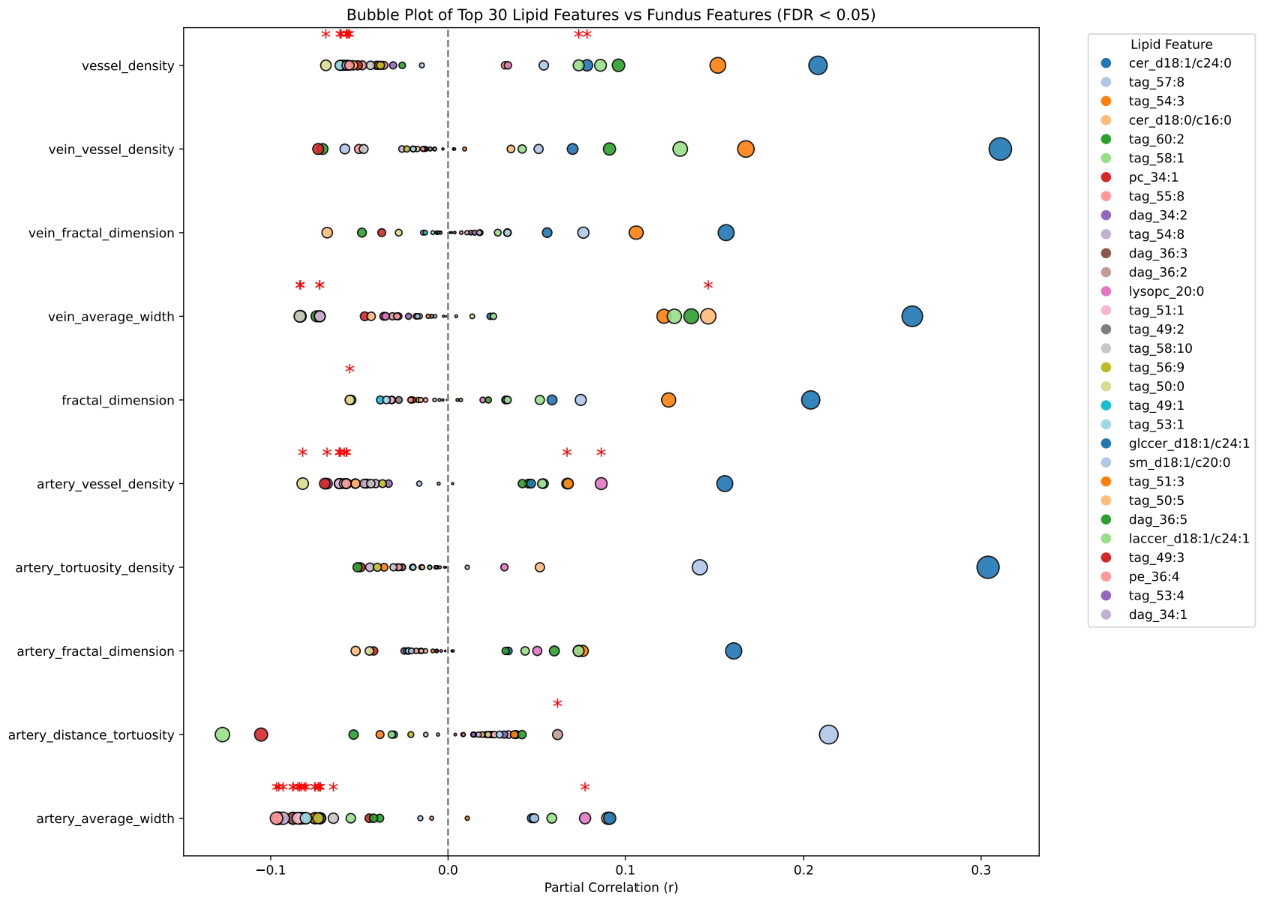}
  \caption{Top 30 Lipid–Retina Associations (Bubble Plot). 
Bubble plot illustrating partial correlations between 30 lipid species and 10 retinal features. Dot size reflects correlation magnitude; red asterisks indicate FDR-adjusted significance.}
  \label{fig:3}
\end{figure}

\begin{figure}[H]
  \centering
  \includegraphics[width=0.9\textwidth]{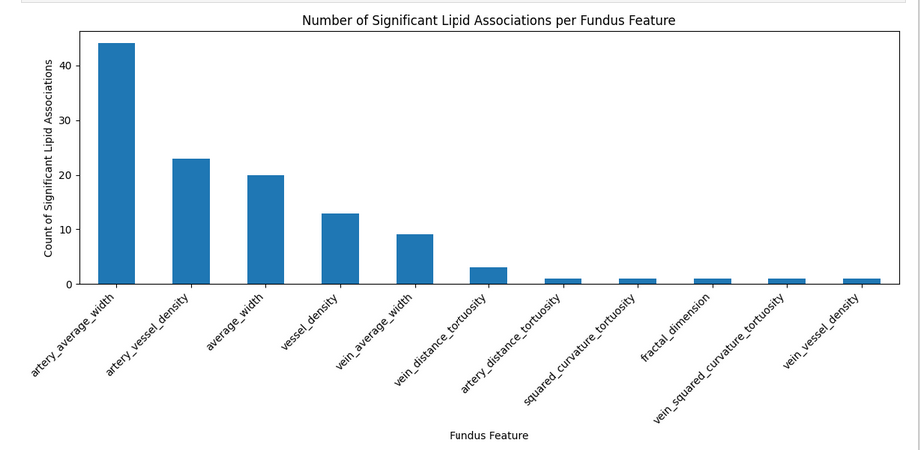}
  \caption{Count of Significant Lipid Associations per Fundus Feature
Bar plot quantifying the number of lipid species significantly associated with each of the fundus features. The artery average width was the most lipid-associated feature.}
  \label{fig:4}
\end{figure}

 \subsection{Lipid–Retina Correlations}
In the early stage, fix the reference values by the retinal vessel for the rest of the study based on age and sex. Building on the existing mechanisms, analyse the linear relationship between lipid traits and retinal attributes, adjusting for partial corrections related to both age and sex for each pair. A comprehensive pairwise analysis was conducted across 18 and 187 features for retinal and lipid traits, respectively. This analysis encompasses the \textit{r} , P-values, 95\% CI, and adjustments made through the FDR multiple correction test. The molecular underpinnings of lipids with the retinal microvascular variation give us an understanding of the association between the serum lipidomics profiles and quantitative fundus features. 

The Fig \ref{fig:3} presents the top 30 lipid features alongside ten microvascular characteristics. Among the six vessels analysed, notable features such as vessel density, artery vessel density, and the average width of arteries and veins demonstrated statistically significant results after multiple corrections, with lipid species having a p-value of less than 0.05. This figure underscores that the most consistent, abundant, and meaningful relationships were identified across the lipid classes of TAGs, DAGs, and Cers. These species are implicated in heart disease, both individually and collectively, due to their roles in lipid metabolism, vascular alterations, and inflammatory processes that contribute to atherosclerosis and cardiovascular dysfunction \cite{bib18-meikle2011plasma,bib19-laaksonen2016plasma}.

All TAG and DAG profiles exhibited a similar association pattern with artery average width and artery vessel density (indicated by an asterisk in Fig \ref{fig:3}), both of which exhibited declining trends in opposite directions. Key TAGs included tag\_50:0, tag\_49:2, and tag\_53:1, along with DAGs such as dag\_36:2 and dag\_36:3. Vessel density, vein average width, artery distance tortuosity, and fractal dimension displayed varying trends; some showed decreases while others overlapped. These metrics exhibited both positive and negative correlations, revealing diverse relationships. Artery tortuosity density and fractal dimension exhibited both positive and negative correlations with TAGs accordingly. Vein average width and vessel density were associated with TAGs, DAGs, Cers, PC, PE, FFA, LACCER, and GLCCER in both directions. Additionally, the other metrics—namely tortuosity density, squared curvature, and distance tortuosity—demonstrated relationships involving both veins and arteries.

\subsection{Discussion}
This section presents an integrated discussion of analytical findings by layering visual representations, biological interpretations, and clinical relevance. The associations between serum lipidomic profiles and retinal microvascular features are explored from multiple angles to derive both statistical insights and translational value. We begin with a visual overview of the most significant associations, followed by interpretation of biological pathways, and end with implications for the assessment of cardiovascular risk.

\begin{figure}[!htb]
  \centering
  \includegraphics[width=1.0\textwidth]{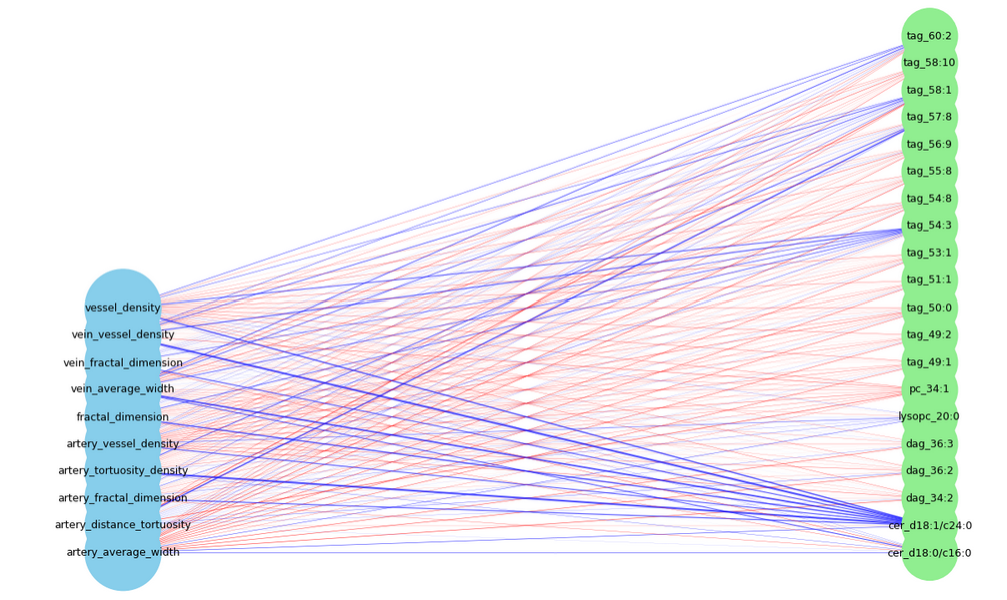}
  \caption{Global Lipid–Fundus Association Network
Network graph linking lipid species (right) to retinal features (left). Node connections indicate direction and strength of partial correlations (red = negative, blue = positive). }
  \label{fig:5}
\end{figure}

\begin{figure}[H]
  \centering
  \includegraphics[width=1.0\textwidth]{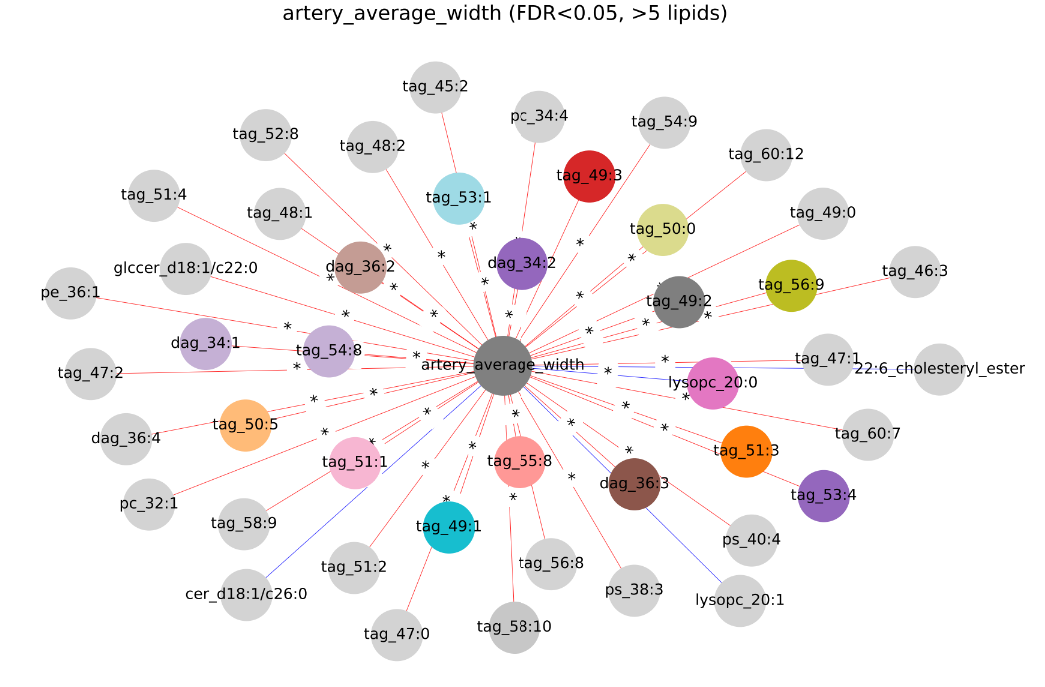}
  \caption{Network diagram illustrating significant associations between artery average width and lipid species (FDR $<$ 0.05, n $>$ 5). The central node represents the retinal feature, while surrounding nodes represent individual lipids. Node colours indicate lipid categories (e.g., TAGs, DAGs, lysophospholipids). Red edges represent negative correlations, blue edges indicate positive correlations, and asterisks mark statistically significant associations. This figure highlights the strong lipid connectivity of the artery average width, particularly with TAG and DAG subclasses. }
  \label{fig:6}
\end{figure}

\begin{figure}[!htb]
  \centering
  \includegraphics[width=0.80\textwidth]{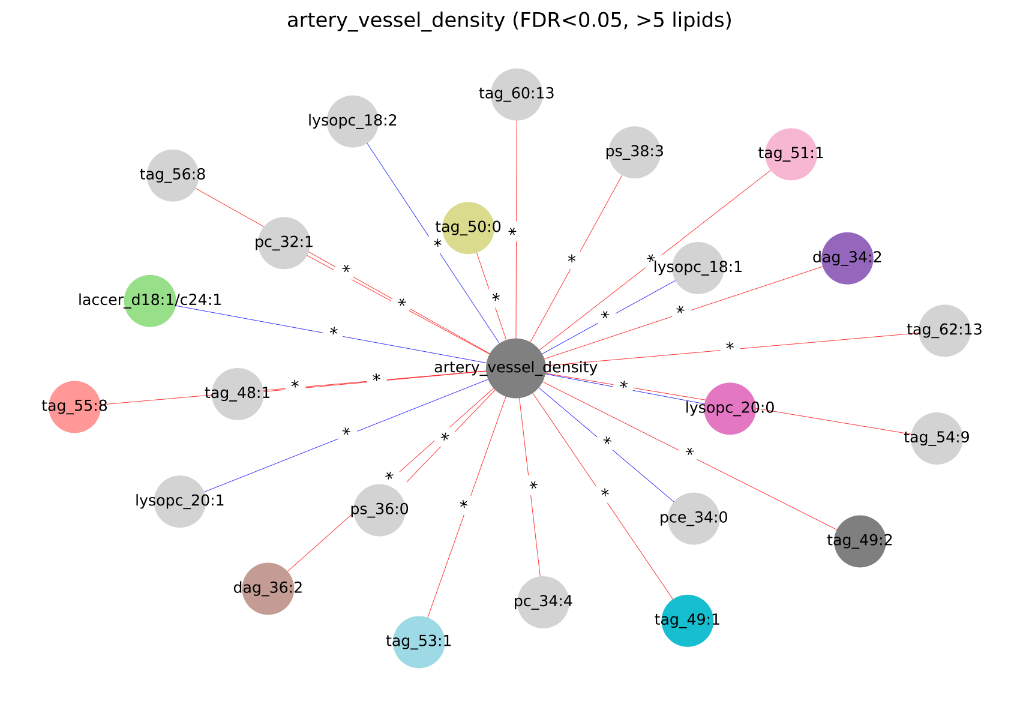}
  \caption{Lipid–artery vessel density network showing significant associations (FDR $<$ 0.05). Node colours reflect lipid classes; red and blue edges indicate negative and positive correlations, respectively. Asterisks mark statistically significant links. }
  \label{fig:7}
\end{figure}

\begin{figure}[H]
  \centering
  \includegraphics[width=0.80\textwidth]{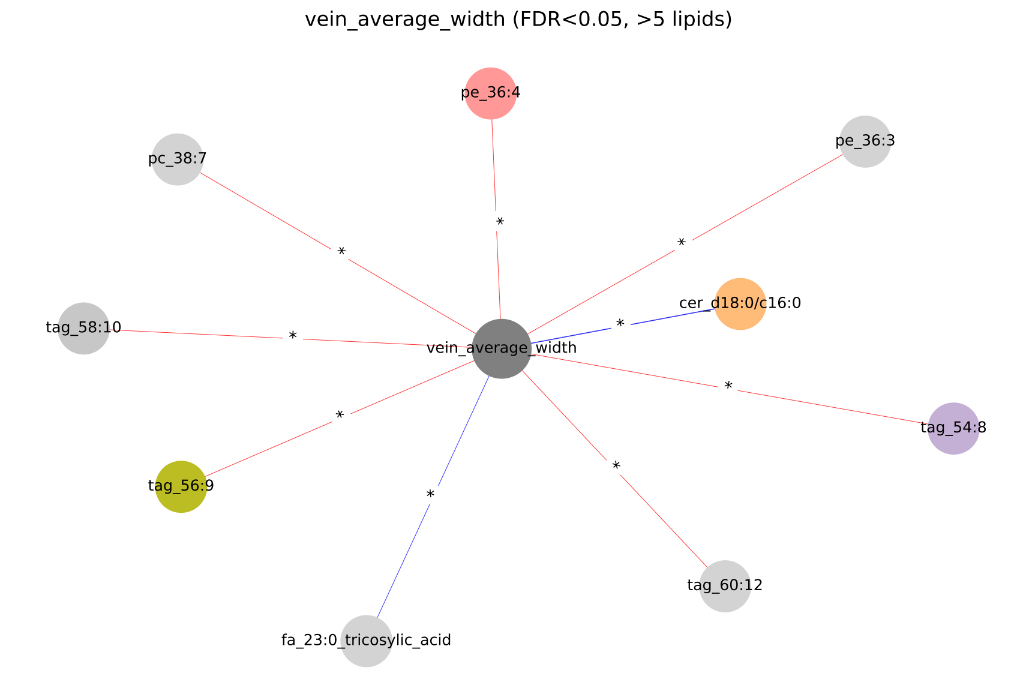}
  \caption{Network of significant lipids associated with vein average width (FDR $<$ 0.05). Node colours indicate lipid types; blue and red edges represent positive and negative correlations. Asterisks highlight significant links. }
  \label{fig:8}
\end{figure}

\begin{figure}[!htb]
  \centering
  \includegraphics[width=0.80\textwidth]{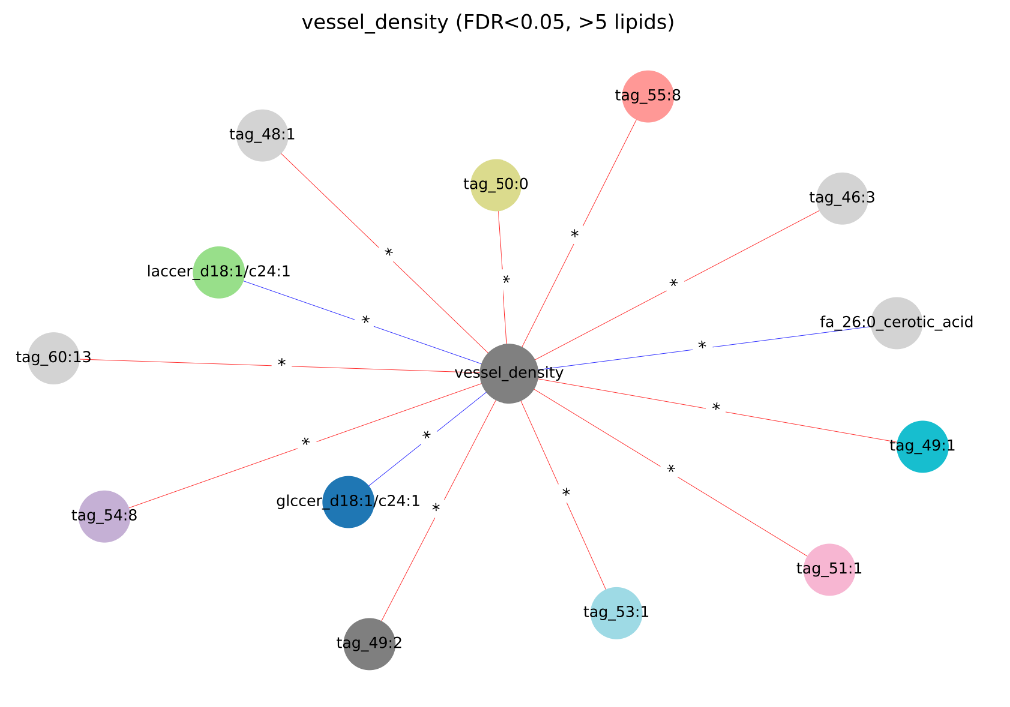}\caption{Network graph showing significant associations between vessel density and lipid species (FDR $<$0.05, $>$5 lipids). Lipids are colored by class, with red edges indicating negative correlations and blue edges indicating positive correlations. The central node represents vessel density, and asterisks denote statistically significant links. }
  \label{fig:9}
\end{figure}

\begin{figure}[!htb]
  \centering
  \includegraphics[width=0.85\textwidth]{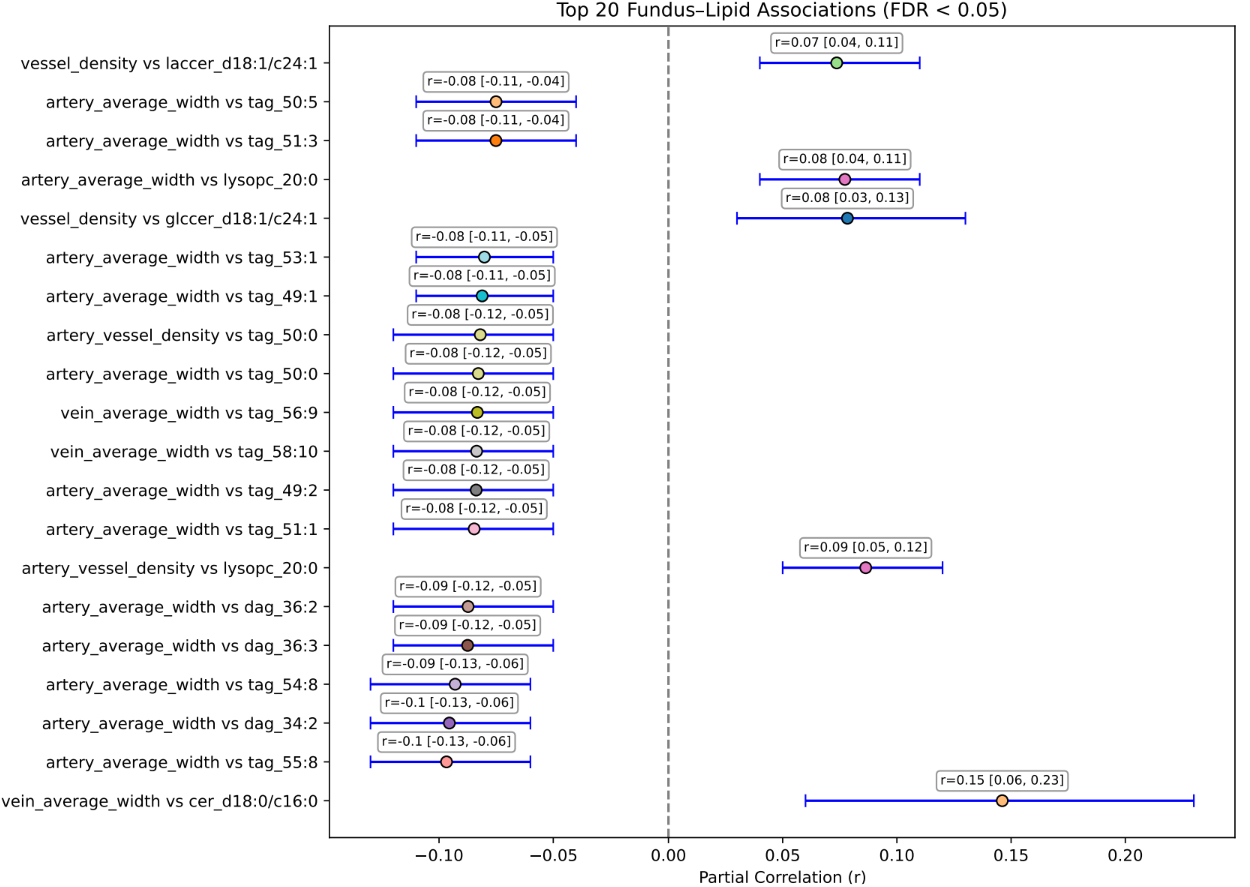}
  \caption{Top 20 Significant Fundus–Lipid Associations (Forest Plot). 
Forest plot showing the strongest associations based on partial correlation magnitude and 95\% confidence intervals. Artery-based features dominate negative correlations with TAGs and DAGs.}
  \label{fig:10}
\end{figure}

\subsubsection{ Visualization of Associations}

To further elucidate and communicate the identified lipid–retina associations, a multiple visual analytics approach was employed. This included bubble plots, network graphs, and ranked correlation bar plots. These visualisations provided not only statistical clarity but also biological insight into the structure and intensity of the observed associations.

The bubble plot in Fig \ref{fig:3} highlights the partial correlations between the top 30 lipid species and ten microvascular features, with dot sizes indicating the strength of correlation and red asterisks marking statistically significant relationships (FDR $<$ 0.05). This visualisation underscores the dominance of arterial features—particularly artery average width, artery vessel density, and vessel density—as major hubs of lipid sensitivity, further simplified in the Figs \ref{fig:5} to \ref{fig:9}. The plot shows a dense clustering of negative correlations for artery calibre and branching density with lipids such as TAGs and DAGs, many of which have been implicated in subclinical vascular injury and atherosclerotic risk \cite{bib31-toledo2017plasma}

A count for each lipid is provided in Fig. \ref{fig:4}, which quantifies the number of significant lipid associations per fundus trait. Here, the artery average width was again identified as the most frequently associated feature, with over 40 significant correlations, followed by artery vessel density and total vessel density. This ranking confirms that vessel structure—particularly within the arterial tree—is more metabolically sensitive than fractal complexity or tortuosity. The finding aligns with prior retinal imaging studies associated with dyslipidemia, which have demonstrated that arterial diameter and branching density in the retina are modifiable in response to systemic lipid levels and cardiometabolic burden \cite{bib32-drobnjak2016retinal}.

To visualise the complexity and specificity of these interactions, Figs. \ref{fig:5} to \ref{fig:9} present shared network diagrams. In Fig. \ref{fig:5}, a global network of significant fundus–lipid associations is displayed, with fundus features positioned on the left and lipids on the right. The lines are used to indicate the direction and strength of the association, colored by the correlation sign. This network representation allows for the clustering of lipids around the artery average width and artery vessel density to be particularly evident. Sub-networks in the subsequent figures from the \ref{fig:6} to \ref{fig:9} provide focused graphs for fundus traits that have more than five lipid associations. For example, in Fig. \ref{fig:6}, the artery average width is shown to be connected to over 20 lipid species, many of which are identified as pro-atherogenic TAGs and DAGs. Similarly, Fig. \ref{fig:7},\ref{fig:8}, \ref{fig:9} details analogous lipid subnetworks for artery vessel density, vessel density, and vein average width.

Finally, Fig. \ref{fig:10} summarises the top 20 significant fundus–lipid associations, arranged by magnitude of partial correlation and adjusted for multiple comparisons. The left side of the plot illustrates strong inverse correlations—particularly between artery average width and several TAG and DAG species (e.g., tag\_55:8, dag\_36:2, tag\_50:0)—while the right side highlights a few positive correlations, such as between vein average width and cer\_d18:0/c16:0 (r = 0.15, 95\% CI: 0.06 to 0.23). This combination of visualisation and statistical annotation facilitates interpretation of both effect size and confidence, aiding downstream clinical translation and hypothesis generation.

Importantly, these visualisations align with and extend the insights gleaned from epidemiological and mechanistic studies. Lipid classes such as ceramides and long-chain TAGs have been linked to increased risk of myocardial infarction, stroke, and vascular dementia \cite{bib19-laaksonen2016plasma,bib31-toledo2017plasma}. Their association with retinal vessel narrowing and reduced perfusion density suggests a convergent biological mechanism involving microvascular dysfunction.

Given the retina's embryological and physiological similarity to the cerebral and coronary microvasculature, these findings reinforce the role of the eye as a window to systemic vascular health. To sum up, a clear prioritisation of biologically plausible lipid-retina interactions was enabled by the visualisation pipeline. These insights, when interpreted alongside previous CVD research, support the potential utility of fundus imaging in identifying lipid-related vascular risks, particularly in asymptomatic or preclinical populations. The average width of arteries was identified as a key biomarker that can capture a range of pivotal lipid serum profiles. This is expected to serve as a non-invasive marker for predicting and diagnosing patients with elevated cardiometabolic risk factors in future studies.

\begin{table}[htbp]
\centering
\caption{Significant correlations of the study population}
\label{tab:significant_correlations}
\begin{tabular}{lllccc}
\toprule
fundus\_feature & lipid\_feature & r & CI\_lower & CI\_upper & P-value \\
\midrule
artery\_average\_width & 22:6\_cholesteryl\_ester &    0.06 &  0.02 &  0.09 & 0.0336 \\
artery\_average\_width & cer\_d18:1/c26:0 &  0.06 &  0.02 &  0.09 & 0.048 \\
artery\_average\_width & dag\_34:1 & -0.07 & -0.11 & -0.03 & 0.0145 \\
artery\_average\_width & dag\_34:2 & -0.10 & -0.13 & -0.06 & 0.0006 \\
artery\_average\_width & dag\_36:2 & -0.09 & -0.12 & -0.05 & 0.0004 \\
artery\_average\_width & dag\_36:3 & -0.09 & -0.12 & -0.05 & 0.0004 \\
artery\_average\_width & glccer\_d18:1/c16:0 & -0.06 &  -0.10 & -0.03 & 0.0221 \\
artery\_average\_width & lysopc\_20:0 & 0.08 & 0.04 & 0.11 & 0.0032 \\
artery\_average\_width & ysopc\_20:1 & 0.06 &  0.03  & 0.09 & 0.0240 \\
artery\_average\_width & pc\_32:1 & -0.06 &  -0.10 & -0.03 & 0.0203 \\
artery\_average\_width & pc\_34:4 & -0.06 &  -0.10 & -0.03 & 0.0126 \\
artery\_average\_width & pe\_36:1 & -0.05 &  -0.09 & -0.02 & 0.0485 \\
artery\_average\_width & ps\_38:3 & -0.07 & -0.10 & -0.04 & 0.0062 \\
artery\_average\_width & tag\_47:1 & -0.07 & -0.11 & -0.04 & 0.0062 \\
artery\_average\_width & tag\_48:1 & -0.07 & -0.10 & -0.03 & 0.0114 \\
artery\_average\_width & tag\_49:1 & -0.08 & -0.11 & -0.05 & 0.0007 \\
artery\_average\_width & tag\_49:2 & -0.08 & -0.12 & -0.05 & 0.0006 \\
artery\_average\_width & tag\_49:3 & -0.07 & -0.11 & -0.04 & 0.0042 \\
artery\_average\_width & tag\_50:0 & -0.08 & -0.12 & -0.05 & 0.0007 \\
artery\_average\_width & tag\_50:5 & -0.08 & -0.11 & -0.04 & 0.0030 \\
artery\_average\_width & tag\_51:1 & -0.08 & -0.12 & -0.05 & 0.0006 \\
artery\_average\_width & tag\_51:3 & -0.08 & -0.11 & -0.04 & 0.0061 \\
artery\_average\_width & tag\_53:1 & -0.08 & -0.11 & -0.05 & 0.0012 \\
artery\_average\_width & tag\_53:4 & -0.07 & -0.11 & -0.04 & 0.0047 \\
artery\_average\_width & tag\_54:8 & -0.09 & -0.13 & -0.06 & 0.0004 \\
artery\_average\_width & tag\_55:8 & -0.10 & -0.13 & -0.06 & 0.0001 \\
artery\_average\_width & tag\_56:8 & -0.07 & -0.11 & -0.04 & 0.0047 \\
artery\_average\_width & tag\_56:9 & -0.07 & -0.11 & -0.04 & 0.0041 \\
artery\_average\_width & tag\_58:10 & -0.06 & -0.10 & -0.03 & 0.0165 \\
artery\_average\_width & tag\_60:7 & -0.07 & -0.11 & -0.03 & 0.0170 \\
artery\_distance\_tortuosity & dag\_36:2 & 0.06 & 0.03 & 0.10 & 0.0223 \\
artery\_vessel\_density & dag\_34:2 & -0.07 & -0.11 & -0.03 & 0.0264 \\
artery\_vessel\_density & dag\_36:2 & -0.06 & -0.09 & -0.03 & 0.0245 \\
artery\_vessel\_density & laccerc\_d18:1/c24:1 & 0.07 & 0.03 & 0.10 & 0.0203 \\
artery\_vessel\_density & lysopc\_18:1 & 0.07 & 0.04 & 0.11 & 0.0082 \\
artery\_vessel\_density & lysopc\_20:0 & 0.09 & 0.05 & 0.12 & 0.0006 \\
artery\_vessel\_density & ps\_38:3 & -0.07 & -0.10 & -0.03 & 0.0078 \\
artery\_vessel\_density & tag\_48:1 & -0.06 & -0.10 & -0.03 & 0.0221 \\
artery\_vessel\_density & tag\_49:1 & -0.06 & -0.10 & -0.03 & 0.0222 \\
artery\_vessel\_density & tag\_49:2 & -0.06 & -0.09 & -0.02 & 0.0368 \\
artery\_vessel\_density & tag\_50:0 & -0.08 & -0.12 & -0.05 & 0.0007 \\
artery\_vessel\_density & tag\_51:1 & -0.06 & -0.10 & -0.03 & 0.0245 \\
artery\_vessel\_density & tag\_53:1 & -0.06 & -0.09 & -0.02 & 0.0335 \\
artery\_vessel\_density & tag\_55:8 & -0.06 & -0.09 & -0.02 & 0.0317 \\
artery\_vessel\_density & tag\_56:8 & -0.06 & -0.09 & -0.02 & 0.0442 \\
fractal\_dimension & tag\_50:0 & -0.06 & -0.09 & -0.02 & 0.0482 \\
squared\_curvature\_tortuosity & pc\_44:12 & -0.07 & -0.10 & -0.03 & 0.0141 \\
vein\_average\_width & cer\_d18:0/c16:0 & 0.15 & 0.06 & 0.23 & 0.0226 \\
vein\_average\_width & pc\_38:7 & -0.07 & -0.10 & -0.04 & 0.0064 \\
vein\_average\_width & pe\_36:4 & -0.07 & -0.11 & -0.04 & 0.0036 \\
vein\_average\_width & tag\_54:8 & -0.07 & -0.11 & -0.04 & 0.0066 \\
vein\_average\_width & tag\_56:9 & -0.08 & -0.12 & -0.05 & 0.0006 \\
vein\_average\_width & tag\_58:10 & -0.08 & -0.12 & -0.05 & 0.0006 \\
vessel\_density & glccer\_d18:1/c24:1 & 0.08 & 0.03 & 0.13 & 0.0428 \\
vessel\_density & laccerc\_d18:1/c24:1 & 0.07 & 0.04 & 0.11 & 0.0086 \\
vessel\_density & tag\_48:1 & -0.06 & -0.09 & -0.02 & 0.0301 \\
vessel\_density & tag\_49:1 & -0.06 & -0.09 & -0.03 & 0.0239 \\
vessel\_density & tag\_49:2 & -0.06 & -0.09 & -0.02 & 0.0365 \\
vessel\_density & tag\_50:0 & -0.07 & -0.10 & -0.03 & 0.0087 \\
vessel\_density & tag\_51:1 & -0.06 & -0.09 & -0.02 & 0.0447 \\
vessel\_density & tag\_53:1 & -0.06 & -0.10 & -0.03 & 0.0245 \\
vessel\_density & tag\_54:8 & -0.06 & -0.09 & -0.02 & 0.0460 \\
vessel\_density & tag\_55:8 & -0.06 & -0.09 & -0.02 & 0.0400 \\
\botrule
\end{tabular}
\begin{tablenotes}
\footnotesize
\item \textbf{Note:} Table shows statistically significant correlations between retinal microvascular features and serum lipid species in a healthy adult cohort, after adjusting fundus traits for age and sex. Only associations meeting the significance threshold (FDR-adjusted  P$<$ $0.05$ ) and recurring across multiple retinal features are reported.
\end{tablenotes}
\end{table}

\clearpage

\subsubsection{Key Biological Interpretations}

The key associations identified across healthy individuals between serum lipidomic profiles and retinal vasculometry were distilled into simplified patterns, revealing subtle motifs that may contribute to age-related disease risk. As summarised in Fig~\ref{fig:5}, \textit{artery average width} emerged as the most lipid-associated retinal feature, with over 40 statistically significant correlations. It was followed by \textit{artery vessel density} ($n \approx 20$) and \textit{vessel density} ($n \approx 10$). Together, these three features accounted for the majority of lipid-related signals, highlighting the sensitivity of vascular calibre and perfusion metrics to circulating lipid composition.

Notably, artery average width captured the largest number of lipidomic associations in this study, and its biological relevance is further supported by its observed decline with increasing age, an effect that was more pronounced in females than in males. This sex-specific pattern of vascular narrowing aligns with established literature on microvascular ageing and hormonal influence on vascular tone. For instance, lipid species such as dag\_34:1, dag\_34:2,pe\_36:1, pc\_32:1, tag\_51:1 and tag\_53:1 have previously been linked to coronary artery disease and cardiovascular events \cite{bib31-toledo2017plasma,bib31a-liu2022multi,bib31b-zeng2022lipidomic}. Elevated plasma lipid levels are known to contribute to endothelial dysfunction, arterial plaque formation \cite{bib15-meikle2014lipidomics}, and an increased risk of heart failure, particularly in postmenopausal women\cite{bib33-wang2018metabolic}, where lipid metabolism and vascular structure undergo significant changes \cite{bib33a-beyene2020high}.

Emerging evidence also suggests that serum lipid profiles can serve as biomarkers of biological ageing \cite{bib33-wang2018metabolic,bib29-reicher2024phenome}. Within this context, our findings may represent an extension of this concept, whereby retinal microvascular features—especially artery average width—may act as a non-invasive readout of lipid-mediated vascular ageing. Accordingly, the identified lipid profiles may serve as surrogate biomarkers for the early detection of subclinical cardiovascular disease through retinal imaging.

Subsequently, arterial features showed stronger associations with lipid metabolism than venous features. For example, artery traits, namely average width and vessel density, were more strongly linked to lipid levels, while vein traits, such as average width and tortuosity, had fewer associations. This may be because arteries are more active in controlling blood pressure and are more likely to change in response to lipid-related stress control~\cite{bib34-ding2025computational}.

Detailed network representations in Figs~\ref{fig:4} to~\ref {fig:9} further contextualise these findings by showing not only the number of associations per feature, but also the nature of the lipid species involved. \textit{Artery average width} was strongly linked to a wide range of lipids, including long-chain TAGs (e.g., \texttt{tag\_48:1}, \texttt{tag\_50:0}, \texttt{tag\_58:10}, \texttt{tag\_50:5}, \texttt{tag\_54:8}), DAGs (e.g., \texttt{dag\_36:2}, \texttt{dag\_34:2}), and lysophospholipids (e.g., \texttt{lysoPC\_20:0}). These lipid species are not only abundant in plasma but have also been implicated in key cardiometabolic processes, such as lipid transport, insulin resistance, and oxidative stress~\cite{bib35-razquin2018plasma,bib36-alshehry2016plasma}. Their consistent inverse association with the caliber of the arteries suggests a pathophysiological mechanism by which elevated lipid levels contribute to microvascular narrowing or rarefaction, a phenomenon known to precede overt cardiovascular disease.

Similarly,  \textit{artery vessel density} was associated with a comparable lipidomic signature, including glycosphingolipids, glycerolipid, and glycerophospholipids such as \texttt{laccer\_d18:1/c24:1}, \texttt{tag\_49:1}, \texttt{tag\_50:0}, {lysopc\_18:1}, and ps\_38:3. These lipids are often enriched in atherogenic lipoprotein particles and have been linked to endothelial apoptosis, vascular inflammation, and arterial stiffening \cite{bib17-chen2023longitudinal} The observed reduction in retinal arterial density with elevated lipid levels may reflect capillary rarefaction and microvascular dysfunction, mechanisms commonly implicated in hypertension and heart failure pathophysiology ~\cite{bib19-laaksonen2016plasma,bib36a-chen2024association}.

A closer look at the top 20 fundus--lipid associations (see Fig~\ref{fig:10}) revealed that many of the strongest correlations involved \textit{artery average width}, which consistently showed negative relationships with saturated and mono-unsaturated TAGs and DAGs. For instance, \texttt{tag\_50:5} ($r = -0.08$, CI: $-0.11$ to $-0.04$), \texttt{dag\_36:2} ($r = -0.09$, CI: $-0.12$ to $-0.05$), and \texttt{tag\_55:8} ($r = -0.10$, CI: $-0.13$ to $-0.06$) were among the most robust associations. These lipids are components of glycerolipid subclasses and have been previously linked to subclinical atherosclerosis and cardiovascular event prediction beyond traditional cholesterol markers~\cite{bib16-stegemann2014lipidomics}. Their consistent inverse relationships with microvascular width suggest a lipid-driven arteriolar constriction effect that may reflect early microvascular damage or metabolic stress.

While fewer associations were observed for venous features, some meaningful patterns emerged. For instance, \textit{vein average width} was positively associated with \texttt{cer\_d18:0/c16:0} ($r = 0.15$, CI: $0.06$ to $0.23$), a ceramide previously linked to cardiometabolic risk and vascular stiffness~\cite{bib19-laaksonen2016plasma}. This relationship may suggest venous dilation as a compensatory response to ceramide-induced perfusion deficits or increased vascular resistance. 

Additionally, \texttt{vessel density }, when considered as a composite measure, also displayed consistent associations with  TAG species and Cers species to the glccer and laccer, further reinforcing the idea that retinal perfusion is highly sensitive to systemic lipid environments. Recent longitudinal lipidomic studies have shown that ceramides, ether-linked lipids, and saturated TAG subclasses dynamically change during aging, infection, and insulin resistance, conditions strongly implicated in microvascular dysfunction\cite{bib38-hornburg2023dynamic}. 
\subsubsection{Clinical and Translational Relevance}

Across all visual and statistical layers of the analysis—including partial correlations, forest plots, bubble charts, and network topologies—a consistent pattern emerged linking specific lipid subclasses to retinal microvascular traits. Three major lipid categories dominated these associations: glycerolipids (e.g., tag\_48:1, tag\_49:1, Tag\_50: 0, tag\_51:1, tag\_53:1,tag\_55:8, dag\_34: 2, dag\_36: 2), sphingolipids (e.g., Cer\_d18:1/c24: 0, Cer d18:0 / c16: 0), and glycerophospholipids (e.g. lysoPC\_18: 2, lysoPC\_20: 0, ps\_38:3). These species appeared repeatedly in connection with arterial characteristics of average width, vessel density, along with the others retinal features - suggesting a shared molecular axis underlying microvascular remodeling, and previous research consistently counts these associations with the outcome and mortality of CVD\cite{bib18-meikle2011plasma,bib39-seah2020plasma,bib17-chen2023longitudinal,bib31b-zeng2022lipidomic}. 

Collectively, these findings highlight a set of core lipid biomarkers—particularly TAGs, DAGs, and ceramides—that exhibit the strongest associations with retinal vascular morphology, particularly within the arterial tree. Their consistent appearance across multiple analytic views suggests they are not only statistical correlates but also biologically meaningful indicators of microvascular remodeling. These insights support the potential of the retina as a non-invasive platform for metabolic risk profiling, advancing the translational value of oculomics in cardiovascular health.

Importantly, while our results support prior knowledge of lipid-mediated vascular stress, they also expand this understanding through a population-scale, multimodal lens. By mapping the most responsive fundus features to the most connected lipid species, this study offers a biologically interpretable and statistically robust framework for future biomarker discovery and translational cardiovascular research. These findings suggest that fundus imaging, when combined with serum lipid profiling, may enable early, scalable CVD screening across asymptomatic populations in a non-invasive manner.

While novel insights into lipid–retina associations in a healthy population are provided by these findings, several limitations are acknowledged. First, the ability to infer causality or determine the directionality of the observed relationships is restricted by the cross-sectional nature of the dataset. Longitudinal follow-up will be deemed essential to establish whether lipidomic changes are preceded by retinal vascular remodeling or result from it.

Second, although age and sex are adjusted for using partial correlation, other covariate variables such as diet, physical activity, medication use (e.g., statins), and alcohol consumption are not available and may have influenced the associations observed. It is suggested that future analyses are conducted to incorporate these covariates through multivariable models or propensity score matching to strengthen causal interpretation.

Third, the analysis has focused on pairwise correlations, which are interpretable but inherently limited in capturing the complex multivariate nature of systemic physiology. Advanced integrative approaches such as multi-task learning, latent factor modeling, or graph-based inference are recommended for uncovering nonlinear or network-level associations in future work.

Finally, a promising framework for supervised prediction of cardiometabolic outcomes is offered by the lipid–fundus patterns revealed here. It is anticipated that this platform will be extended through external validation, multi-timepoint modeling, and explainable machine learning methods, ultimately contributing to non-invasive risk stratification strategies in cardiovascular health.

To conclude, this study presents the first population-scale analysis linking deep-learning-derived retinal microvascular features with detailed lipidomic profiles in healthy individuals. The findings suggest that arterial features—particularly artery average width—are robustly associated with specific lipid subclasses, including TAGs, DAGs, and ceramides. These results underscore the potential of fundus imaging as a non-invasive platform for early cardiometabolic risk stratification and highlight the need for longitudinal and multi-modal integration in future research.






\begin{figure}[H]
  \centering
  \includegraphics[width=1.0\textwidth]{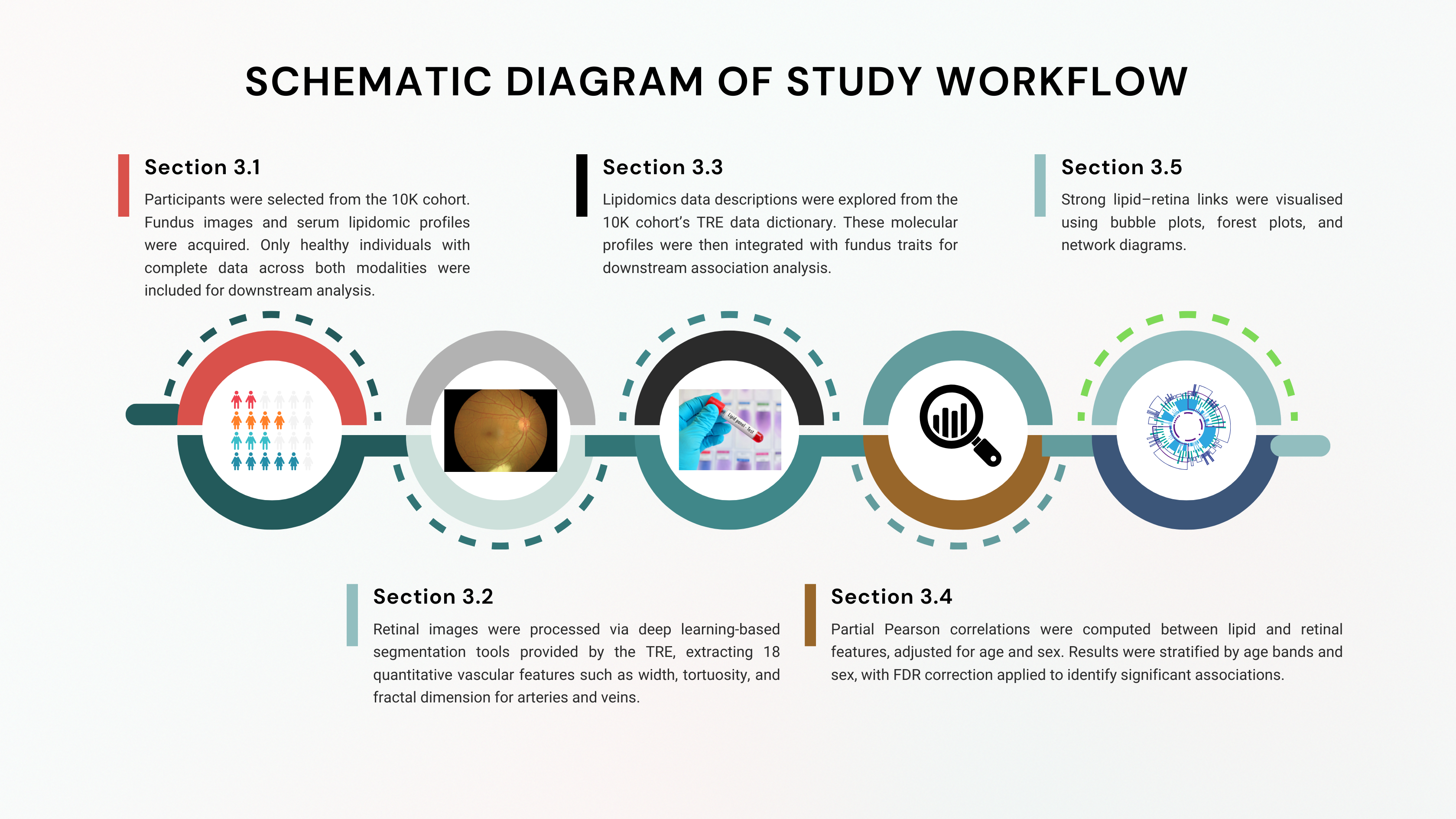}
  \caption{Study Workflow Diagram
Schematic showing data acquisition, preprocessing, demographic adjustment, correlation analysis, and visualization stages of the retinal–lipidomics study. }
  \label{fig:11}
\end{figure}

\section{Methods}\label{sec11}
The research overview and the work mechanism are displayed in the schematic diagram in Fig \ref{fig:11}. The study went through the process of data acquisition, pre-processing, merging, analysis, biological implications, and interpretation. This research is conducted within the framework of a trusted research environment (TRE), designed to facilitate secure access for the researcher and collaborators in pursuit of the research objectives. The two studies, fundus and serum lipidomics, are acquired for downstream analysis. 

\subsection{Study Population and Data}

The human phenotype project (HPP) is a longitudinal cohort study associated with a large biobank of more than 13,000 individuals who have completed their initial visit. The initial goals set by the 10k cohort were to investigate variations in disease susceptibility, clinical phenotypes, and therapeutic responses between different
individuals, using the 10k cohort study, to make a connection between genotypes, phenotypes, and eliminate the complex interaction between them. This will help to develop risk prediction models for chronic type 2 diabetes and CVD, and obesity clinical outcomes for a timely intervention. Subsequently, with respect to this, the scientific community can answer many of the questions \cite{bib40-shilo202110}. The increasing incidence and fatality rate of cardiovascular disease requires a non-invasive way to detect it, other than the traditional way. To counteract this, the proposed work may be a surrogate marker of early-onset diseases. The baseline demographic characteristics and clinical data, namely age, sex, and unique identifiers, were considered a basic part of the 10k cohort.

Upon arrival at the research site, all participants gave their written informed consent in accordance with the principles established in the Declaration of Helsinki. To maintain confidentiality, all identification information was removed prior to data processing. The study protocol received approval from the Institutional Review Board of the Weizmann Institute of Science \cite{bib30-shapira2024unveiling}. In this study, 7,068 and 6,321 adult participants from the retinal microvasculature and serum lipidomics datasets were acquired by passing through the measurements protocols set by the 10k cohorts \cite{bib41-phenoAIfundus,bib42-phenoAIserum}. Bilateral fundus images were acquired using the Icare DRSplus confocal fundus imaging system (ICare), which captures a 45◦ macula-centred field of view without pupil dilation. Fasting venous blood samples were collected for lipidomic profiling and analysed using an untargeted (UHPLC-ESI-HRMS). Lipid identification and quantification were performed using the \url{https://knowledgebase.pheno.ai} curated lipidomics pipeline (data
set 008) \cite{bib42-phenoAIserum}, which applies internal standards, retention time, alignment, and signal normalisation to generate standardised lipid intensity measures suitable for downstream statistical analyses.

\begin{table}[h]
\caption{Baseline characteristics of the study population}\label{tab1}%
\begin{tabular}{@{}llll@{}}
\toprule
Characteristics, mean ± SD or counts \% & Male (n= 3443) (48.7\%) & Female (n= 3625) (51.3\%) & All (n=7068)\\
\midrule
Main Fundus Features \\
\midrule
age    & 52.17 ± 7.84  & 53.08 ± 7.87  & 52.64 ± 7.87  \\
artery\_average\_width    & 18049.7637 ± 1269.9832   & 18549.1074 ± 1255.8112  & 18305.1864 ± 1287.0806 \\
artery\_vessel\_density   & 0.0391 ± 0.0044   & 0.0390 ± 0.0047  & 0.0391 ± 0.0045  \\
artery\_tortuosity\_density    & 0.6929 ± 0.0326   & 0.7010 ± 0.0337  & 0.6971 ± 0.0334)  \\
vein\_average\_width    & 19198.0712 ± 1231.6286   & 19619.0650 ± 1251.8373  & 19413.9884 ± 1259.6486  \\
vein\_vessel\_density   & 0.0500 ± 0.0042   & 0.0498 ± 0.0044  & 0.0499 ± 0.0043  \\
vein\_tortuosity\_density    & 0.7038 ± 0.0246  & 0.7080 ± 0.0250  & 0.7060 ± 0.0249 \\
\midrule
Main Lipids Features & Male (n= 2944) (46.6\%) & Female (n= 3377) (53.4\%)  & All (n=6321)\\
\midrule
age    & 51.26 (8.11)  & 52.19 (8.05)  & 51.76 (8.09)  \\
22:6\_cholesteryl\_ester    & 0.8892 ± 0.3524   & 1.0128 ± 0.3314  & 0.9556 ± 0.3468  \\
cer\_d18:0/c16:0   & 0.3015 ± 0.2206   & 0.2706 ± 0.2132  & 0.2842 ± 0.2169  \\
cer\_d18:1/c24:0    & 1.0004 ± 0.1305   & 1.0030 ± 0.1217  & 1.0017 ± 0.1254  \\
coenzyme\_q10    & 1.0053 ± 0.2026   & 0.9692 ± 0.1991  & 0.9859 ± 0.2015  \\
dag\_32:1 & 0.0477 ± 0.0965   & 0.0290 ± 0.0653  & 0.0374 ± 0.0813  \\
dag\_34:1    & 0.0489 ± 0.0605   & 0.0329 ± 0.0449  & 0.0403 ± 0.0532 \\
dag\_34:2    & 0.1112 ± 0.0581   & 0.0936 ± 0.0491  & 0.1017 ± 0.0541  \\
dag\_36:2   & 0.2468 ± 0.1303   & 0.2050 ± 0.1046  & 0.2243 ± 0.1190  \\
dag\_36:3    & 0.3447 ± 0.1352   & 0.2924 ± 0.1139  & 0.3165 ± 0.1269  \\
dag\_36:4    & 0.2068 ± 0.2140   & 0.1292 ± 0.1638  & 0.1657 ± 0.1930  \\
dag\_36:5 & 0.0477 ± 0.0965   & 0.0290 ± 0.0653  & 0.0374 ± 0.0813  \\
dag\_34:1    & 0.0228 ± 0.0009   & 0.0214 ± 0.0085  & 0.0222 ± 0.0087 \\
fa\_22:0\_behenic\_acid    & 0.1900 ± 0.0707   & 0.1743 ± 0.0583  & 0.1816 ± 0.0648  \\
glccer\_d18:1/c16:0 & 0.2084 ± 0.1182   & 0.2057 ± 0.1192  & 0.0374 ± 0.0813  \\
glccer\_d18:1/c24:1    & 0.3589 ± 0.0926   & 0.3663 ± 0.0968  & 0.3629 ± 0.0950 \\
laccer\_d18:1/c24:1    & 0.2072 ± 0.0791   & 0.2059 ± 0.0836  & 0.2065 ± 0.0815  \\
lysopc\_20:0   & 0.1265 ± 0.0474   & 0.1185 ± 0.0456  & 0.1222 ± 0.0466  \\
lysopc\_18:2    & 0.8892 ± 0.2970   & 0.8614 ± 0.2937  & 0.8742 ± 0.2955  \\
pc\_32:1    & 1.0533 ± 0.5342  & 1.1672 ± 0.5303  & 1.1143 ± 0.5351  \\
pc\_34:1 & 0.4720 ± 0.1919   & 0.4949 ± 0.2046  & 0.4842 ± 0.1991 \\
pc\_36:0    & 0.0006 ± 0.0004   & 0.0007 ± 0.0004  & 0.0007 ± 0.0004 \\
pc\_40:5   & 0.3978 ± 0.1034   & 0.4124 ± 0.1099  & 0.4054 ± 0.1071  \\
pc\_40:8    & 0.9318 ± 0.2193   & 0.9511 ± 0.2258  & 0.9423 ± 0.2230  \\
pc\_42:3    & 0.6125 ± 0.1665  & 0.6544 ± 0.1715  & 0.6350 ± 0.1705  \\
pce\_34:0 & 0.3588 ± 0.0648   & 0.3912 ± 0.0692  & 0.3761 ± 0.0691 \\
pe\_36:3    & 0.4067 ± 0.2063   & 0.4283 ± 0.2085  & 0.4183 ± 0.2077 \\
pe\_36:4    & 0.3241 ± 0.3516   & 0.3190 ± 0.3449  & 0.3214 ± 0.3480 \\
pg\_34:0   & 0.1339 ± 0.0481   & 0.1290 ± 0.0448  & 0.1313 ± 0.0464  \\
pi\_38:5    & 0.8432 ± 0.2075   & 0.8816 ± 0.2198  & 0.8639 ± 0.2151  \\
ps\_36:0    & 0.3939 ± 0.1114  & 0.4301 ± 0.1183  & 0.4133 ± 0.1165  \\
ps\_38:3 & 0.7524 ± 0.1503   & 0.8043 ± 0.1590  & 0.7802 ± 0.1572 \\
ps\_40:4    & 0.9269 ± 0.2256   & 0.9770 ± 0.2274  & 0.9534 ± 0.2279 \\
sm\_d18:1/c20:0   & 0.2463 ± 0.0524   & 0.2526 ± 0.0522  & 0.2496 ± 0.0524  \\
tag\_47:1    & 0.2791 ± 0.2182   & 0.2670 ± 0.2057  & 0.2726 ± 0.2116  \\
tag\_48:1    & 0.1709 ± 0.1053  & 0.1640 ± 0.1047  & 0.1672 ± 0.1050  \\
tag\_49:1 & 0.3275 ± 0.1795   & 0.3017 ± 0.1686  & 0.3136 ± 0.1742 \\
tag\_49:2 & 0.3119 ± 0.1570   & 0.2937 ± 0.1481  & 0.3022 ± 0.1526 \\
tag\_49:3    & 0.3978 ± 0.1756   & 0.3847 ± 0.1689  & 0.3908 ± 0.1721 \\
tag\_50:0    & 0.7341 ± 0.1979   & 0.7064 ± 0.1880  & 0.7192 ± 0.1931 \\
tag\_50:5   & 0.1339 ± 0.0481   & 0.1290 ± 0.0448  & 0.1313 ± 0.0464  \\
tag\_58:10    & 1.1926 ± 0.5455   & 1.1307 ± 0.5354  & 1.1593 ± 0.5409  \\
tag\_60:7    & 0.5966 ± 0.5317  & 0.5833 ± 0.5225  & 0.5894 ± 0.5268  \\
tag\_62:14 & 0.5258 ± 0.5546   & 0.5565 ± 0.5926  & 0.5423 ± 0.5755 \\
\botrule
\end{tabular}
\begin{tablenotes}
\footnotesize
\item \textbf{Note:} Retinal microvascular features were reported in \textbf{pixels} (e.g., vessel width), \textbf{unitless ratios} (e.g., vessel density), or \textbf{unitless scores} (e.g., fractal dimension, tortuosity). These features were computed within the HPP Trusted Research Environment using a standardized image analysis pipeline applied to 45° fundus images acquired with the iCare DRSplus system (1~pixel $\approx$ 4.3~\textmu m assumed)\cite{bib30-shapira2024unveiling}. Lipidomic features were derived from untargeted UHPLC-ESI-HRMS, and values were expressed as log$_{10}$-transformed relative intensities \cite{bib29-reicher2024phenome}. All variables in this table were aggregated, harmonized, and statistically summarized as \textbf{mean ± standard deviation}.
\end{tablenotes}
\end{table}

\clearpage
\subsection{Retinal Image Processing and Feature Extraction}
\label{Retinal Image Processing}
The participants who have reported a diagnosis of ocular pathology, namely, cataracts, glaucoma, and retinal detachments, were excluded from this study. Fundus images were processed using an open-
source automated AutoMorph framework developed by Zhou
et al \cite{bib12}. This framework has the capability to analyse the
fundus retinal features such as image preprocessing, quality
grading, anatomical segmentation, and vascular morphology
measurements. 

A total of 36 retinal microvascular characteristics were computed from each image and categorized into four primary vascular measurement domains: Fractal Dimension (unitless), Vessel Density (unitless ratio), Average Width (in pixels; 1 pixel $\approx$ 4.3~\textmu m), and Tortuosity Metrics. These measurements are consistent with established practices in recent large-scale retinal studies \cite{bib30-shapira2024unveiling}

\begin{itemize}
    \item \textbf{Fractal Dimension (FD):} Quantifies the structural complexity of the retinal vasculature using the box-counting method based on the Minkowski–Bouligand dimension \cite{bib43-avakian2002fractal}.
    \item \textbf{Vessel Density:} Defined as the ratio of the cumulative area occupied by arteries or veins to the total area of the retinal image.
    \item \textbf{Average Width:} Computed by dividing the total number of vessel pixels by the centerline length of the vessels. Reported in pixels.
    \item \textbf{Tortuosity Metrics:} Include three distinct measurements:
    \begin{enumerate}
        \item \textbf{Distance Tortuosity} – a unitless ratio of the actual path length to the straight-line distance,
        \item \textbf{Curvature Tortuosity} – a unitless measure of vessel path curvature normalised by length \cite{bib44-hart1999measurement},
        \item \textbf{Tortuosity Density} – expressed in \textbf{1/pixels}, reflecting local curvature per unit vessel length \cite{bib45-grisan2008novel}
    \end{enumerate}
\end{itemize}
For each participant, both left and right fundus images were analyzed, and corresponding feature values were averaged to yield a single set of \textbf{18 retinal microvascular features} per individual. This resulted in a dataset of 7068 participants × 18 features, used in subsequent statistical analyses. Summary statistics stratified by sex and vessel type are presented in Table \ref{tab1}.

\subsection{Lipidomics Data}
\label{lipidomics Data}
The lipid species were quantified and annotated by the HPP team utilising established reference libraries and enhanced quality control protocols. Further details regarding sample preparation, measurements, and data processing can be found in the provided resources \cite{bib29-reicher2024phenome,bib42-phenoAIserum}. 

The serum profiles were obtained through the TRE and systematically organized into a reference data dictionary, which encompasses distinct lipid subclasses: 22:6\_cholesteryl\_ester, acca\_, Cer\_, coenzyme\ q10, dag\_, fa\_, gsl\_, glccer\_, laccer\_, lysopc\_, pc\_, pe\_, pg\_, pi\_, ps\_, sm\_, tag\_, and lysope\_. Each subclass comprises a range of molecular species, distinguished by unique fatty acyl chain lengths and degrees of unsaturation. For example, within the Cer subclass, species such as Cer(d18:1/16:0), Cer(d18:1/18:0), and Cer(d18:0/16:0), which vary in their sphingoid bases and fatty acid chain compositions, can be found. Similar diversity is observed among the other subclasses.

This analysis was conducted on a large cohort of 7,068 participants, of which 6,321 were healthy adults (3,443 males and 3,625 females) with complete retinal imaging data, and 2,944 males and 3,377 females had available serum lipidomics profiles. The mean age was 52.64 ± 7.87 years for those with fundus imaging and 51.76 ± 8.09 years for the lipidomics group, with comparable age distributions across sexes. 
Retinal microvascular characteristics were extracted using validated automated software, capturing essential morphological traits such as the average width of arteries and veins, vessel density, fractal dimension, and tortuosity metrics. Serum lipidomics analysis was conducted using ultra-high-performance liquid chromatography coupled with high-resolution mass spectrometry, resulting in the quantification of over 187 lipid species, including 22:6 cholesteryl \_ester, cer\_, dag\_,pc\_, tag\ among others.

\subsection{Data Preprocessing, Merging and Statistical Analysis}

The data cast with the basic building block represented in Table \ref{tab1} and explained in \ref{Retinal Image Processing} and \ref{lipidomics Data} were further processed with the covariate adjustments. Given the literature evidence on the impact of demographic factors \cite{bib25-poplin2018prediction,bib21-ikram2006retinal} on retinal vascular morphology and systemic lipid metabolism, the fundus features were adjusted for age and sex prior to the downstream analyses. Partial correlation analyses were conducted to quantify the relationships between fundus traits and age while sex was controlled for, and vice versa. This methodological approach ensured that the subsequent associations with lipid features would reflect relationships independent of these confounding variables.

Fig. \ref{fig:1} and \ref{fig:2}  illustrate these results, providing clear evidence that age-related diseases such as cardiovascular disease tend to progress with advancing age. Furthermore, the impact of CVD-related processes differs between sexes. For instance, hormonal changes during the menopausal transition influence vascular and metabolic pathways in females, leading to differing susceptibilities in premenopausal and postmenopausal women\cite{bib29-reicher2024phenome}. These patterns of disease progression show distinct interactions across age and sex groups \cite{bib33a-beyene2020high}, which are reflected in the relationships between fundus features and lipidomic profiles. These stratified associations are further explored in the Results section.

The microvasculature features of the fundus were obtained utilising a DL-based segmentation implemented through the Automorph Package. These features were quantified in the form of various objective
measurements, including FD, average arterial width, vessel density, and tortuosity, for both arterial and venous structures. These ocular markers further average bilateral eyes and are paired with the 187 cleaned and normalised lipidomics features. The key features of lipidomics include over 180 and 15 average retinal features, along with the unique participant\_id, age and sex. Before proceeding, the 18 fundus features across different ages were initially assessed using Pearson correlation to evaluate their associations, with sex included as a covariate. The Pingouin package was utilised to compute the correlation coefficients. Additionally, a 95\% CI was established, along with the application of the False Discovery Rate in multiple tests to reject the null hypothesis, thereby identifying statistically significant features. Out of the 15 features examined, only those that met these criteria were selected for further analysis. 

Both datasets were merged using participant\_id, age, and sex to prepare for pairwise statistical analysis. Partial correlation analysis was then performed between fundus and lipidomics features, adjusting separately for age and sex. For each feature pair, we calculated the partial correlation coefficient (r), corresponding p-value, and 95\% CI. All p-values were corrected for multiple comparisons using the BH-FDR method, with statistical significance defined as an adjusted p-value $<$ 0.05. The analysis revealed several plausible and statistically significant linear associations between retinal phenotypes and lipid profiles. These findings are elaborated in the Methods section and visually presented in the Results through forest plots, bubble plots, and network graphs.

\subsection{Visualisation and Data Interpretation}

To support the interpretation of high-dimensional associations between retinal microvascular traits and serum lipidomic profiles, a structured visual analytics pipeline implemented in Python was employed.

The approach began with an initial assessment of how fundus features varied by age and sex in a healthy population. Demographic effects on key retinal measurements were explored using Pearson correlation, age-density plots, and sex-specific boxplots. For example, clear declines in artery average width and vessel density were observed with age, while increases in tortuosity were noted. It was also found that wider vessels were exhibited by males than females across several features. These findings highlighted the importance of controlling for age and sex prior to further analysis.

In the next stage, partial correlation analysis was applied using age and sex as covariates. This allowed for the identification of direct associations between fundus traits and lipidomic features without the confounding influence of demographic factors.

To visualize the significant associations (FDR-adjusted p $<$ 0.05), three main formats were used:

\begin{itemize}
    \item Bubble plots were generated to display the strength and direction of the top lipid–retina correlations. Each bubble represented a feature pair, with size indicating the absolute correlation coefficient ($|r|$) and colour denoting the sign (positive or negative).
    \item Network graphs were constructed using the \verb|networkx| library to highlight clustering patterns and central fundus features most connected to lipid species. Key lipid hubs and vascular targets were revealed by these visualizations.
    \item Ranked bar plots displayed the top 20 strongest associations in descending order of effect size, annotated with confidence intervals and statistical significance.
\end{itemize}

Visual stratification was also performed to explore age- and sex-specific differences in retinal traits, providing context for the correlation analysis. While the stratified visuals focused on the fundus features themselves, they informed an understanding of biological variability and justified the need for covariate adjustment.

Since fundus features were first assessed and adjusted for age and sex, all downstream lipidomic analyses were conducted within this adjusted framework. In this sense, lipidomics was investigated under the umbrella of retinal morphology, where the vascular traits reflect demographic-independent biological variation. This approach ensured that meaningful molecular–vascular relationships were captured by the partial correlation results, rather than indirect effects driven by age or sex.

This multi-layered visual analytics strategy allowed for the presentation of the correlation findings in a form that was both statistically rigorous and biologically interpretable, providing a foundation for future predictive modeling and biomarker discovery.

\backmatter




\section*{Data Availability}
The data used in this study are part of the Human Phenotype Project and were accessed through a secure Trusted Research Environment. Due to data privacy policies, individual-level data is not publicly available. However, academic and non-profit researchers may apply for access via the official HPP Knowledgebase at \url{https://knowledgebase.pheno.ai} or \url{https://humanphenotypeproject.org}. All inquiries and access requests should be directed to \texttt{info@pheno.ai}, including a brief project description and institutional affiliation.

\bmhead{Acknowledgements}

We would like to thank the authors of the study "Unveiling Associations Between Retinal Microvascular Architecture and Phenotypes Across Thousands of Healthy Subjects" \cite{bib30-shapira2024unveiling}for providing valuable inspiration for this work. We are especially grateful to Dr. Hagai Rossman for his helpful guidance in understanding key retinal microvascular terms and concepts. While we followed our own research direction, their work served as an important reference point during the development of this study.

\bibliography{sn-bibliography}

\end{document}